\documentclass[letterpaper, 10 pt, conference]{ieeeconf}  

\IEEEoverridecommandlockouts                              

\usepackage{graphicx}
\usepackage{comment}
\usepackage{array}
\usepackage{booktabs}
\usepackage{adjustbox}

\usepackage[table]{xcolor}
\usepackage{xcolor}

\colorlet{shadecolor}{yellow}
\graphicspath{{Transaction/figure_transaction/}{figure_transaction/}}
\DeclareGraphicsExtensions{.pdf,.jpeg,.png,.eps}

\usepackage[cmex10]{amsmath}
\usepackage{array}
\usepackage{mdwmath}
\usepackage{mdwtab}
\usepackage{eqparbox}
\usepackage{url}
\usepackage{longtable}
\usepackage{lipsum}
\usepackage{blindtext}
\usepackage{amsmath}
\usepackage{multirow}
\usepackage[flushleft]{threeparttable}
\usepackage{color}
\usepackage{times}
\usepackage{epsfig}
\usepackage{graphicx}
\usepackage{amsmath}
\usepackage{amssymb}
\usepackage{tabularx}
\usepackage{cite}
\usepackage{makecell}
\usepackage{comment}
\usepackage[normalem]{ulem}
\usepackage{tabularx}
\usepackage{import}
\usepackage{array}
\usepackage{boldline}
\usepackage{caption}
\usepackage{subcaption}

\usepackage{adjustbox}

\usepackage{url}

\usepackage{xurl}

\title{\LARGE \bf
Wearable-based Classification of Running Styles with Deep Learning

}

\author{Setareh Rahimi Taghanaki$^{1}$, Michael Rainbow$^{2}$, Ali Etemad$^{3}$
\thanks{$^{1, 3}$Setareh Rahimi Taghanaki and Ali Etemad are with the Department of Electrical and Computer Engineering,
Queen's University, Kingston, Canada, {\tt\small \{19srt3, ali.etemad\}@queensu.ca}}%
\thanks{$^{2}$Michael Rainbow is with the  the Department of Mechanical Engineering, Queen's University, Kingston, Canada, {\tt\small michael.rainbow@queensu.ca}}%
}

\begin{document}

\maketitle
\thispagestyle{empty}
\pagestyle{empty}

\begin{abstract}
Automatic classification of running styles can enable runners to obtain feedback with the aim of optimizing performance in terms of minimizing energy expenditure, fatigue, and risk of injury. To develop a system capable of classifying running styles using wearables, we collect a dataset from 10 healthy runners performing 8 different pre-defined running styles. Five wearable devices are used to record accelerometer data from different parts of the lower body, namely left and right foot, left and right medial tibia, and lower back. Using the collected dataset, we develop a deep learning solution which consists of a Convolutional Neural Network and Long Short-Term Memory network to first automatically extract effective features, followed by learning temporal relationships. Score-level fusion is used to aggregate the classification results from the different sensors. Experiments show that the proposed model is capable of automatically classifying different running styles in a subject-dependant manner, outperforming several classical machine learning methods (following manual feature extraction) and a convolutional neural network baseline. Moreover, our study finds that subject-independent classification of running styles is considerably more challenging than a subject-dependant scheme, indicating a high level of personalization in such running styles. Finally, we demonstrate that by fine-tuning the model with as few as 5\% subject-specific samples, considerable performance boost is obtained.

\end{abstract}

\section{Introduction}
Running is a common aerobic exercise which many people worldwide partake in as a sport. 
It is often considered one of the most historic types of competitive sport, dating back centuries \cite{hausken2018evolutions}. Despite its pervasiveness, running can be performed in a wide variety of different `\textit{styles}' \cite{goss2012review, goss2012comparison}, often as a result of different physiological/biomechanical factors \cite{ceyssens2019biomechanical}, training \cite{blagrove2018effects}, and even the intention of the runner. These variations can play major roles in energy expenditure \cite{pedersen2007effects}, speed/endurance \cite{vorup2016effect}, and even injury \cite{goss2012review, kozinc2017common, messier1988etiologic}. 

Based on the above, we believe it is highly valuable and beneficial to develop systems capable of automatically detecting different running styles. Such a system can be used for injury analysis, coaching and training, and other applications. Moreover, we believe the use of wearable sensors for such a system, as opposed to cameras for instance, will allow for more detailed, exact, and targeted monitoring as well as lack of occlusions of key body parts.

The notion of using wearables and machine learning to classify different activities has been widely studied in the literature. For instance, in \cite{romera2013one}, hand-crafted features were used along with an ensemble of Support Vector Machine (SVM) learners to classify daily activities using wearable data. Similar approaches have been used in \cite{anguita2012human}, \cite{buber2014discriminative}, and \cite{yang2009toward}, where different classical machine learning solutions have been used. In \cite{kulchyk2019activity}, deep learning, in the form of a Convolutional Neural Network (CNN) was used to perform end-to-end classification of different activities. In \cite{mutegeki2020cnn}, the learned representations from a CNN were fed to a Long Short-Term Memory (LSTM) network to classify actions. More recently, self-supervised solutions have been proposed in \cite{saeed2019multi}, \cite{haresamudram2020masked}, \cite{saeed2020federated}, and \cite{rahimi2020self} to reduce reliance on labels and learn more generalized representations by means of pre-text training with a number of transformations applied to wearable data. However, despite the rich available literature on classification of different activity classes, the notion of detecting specific running styles with machine learning, which may be manifested as a result of very minor and subtle changes in running patterns, has not yet been explored.

To tackle the problem of automatic wearable-based running style classification, we design a system for automatic wearable-based classification of different running styles with deep learning. We first record a rich dataset of different subjects running in a number of different pre-defined running styles. The data are recorded using several wearable sensors placed strategically on different parts of the body including each foot, each medial tibia, and on the lower back. We then use this dataset to train deep neural networks for classification of the different styles. Our results show that our proposed CNN can accurately classify the different running styles and outperforms some classical machine learning models, namely support vector machine (SVM), naive Baysed (NB), decision tree (DT), random forest (RF). 
Interestingly, however, we find that the performance considerably drops when the experiments are performed in a leave-one-subject-out manner, even when different CNN architectures are explored. This finding points to the fact that running styles, even in pre-defined categories, are highly subject-specific and vary across different runners. 

Lastly, we demonstrate that by fine-tuning the model with small portions of unseen subjects' samples, a more robust performance is achieved.

\section{Methodology}

\subsection{Dataset}
10 healthy and unimpaired runners (5 women and 5 men), ranging in age between 20 to 26 years, were recruited for this study \cite{rodgers2020investigating}. One of the inclusion criteria was for participants to have been running around 10 km per week over the last three months leading to the data collection session. For each subject, a set of 5 IMeasureU \cite{ImeasureU} Inertial Measurement Units (IMUs) sensors were placed on each foot (top of the shoe, attached through laces), each medial tibia (attached with elastic Velcro straps), and one at the centre of mass on the lower back (attached with Velcro to a torso wrap) to record the accelerations and angular changes of their body segments when running. The accelerometer data are sampled at 500 Hz and collected by a personal computer via Bluetooth. For each subject, their height, weight, age, and sex were also recorded. For all the experiments, an instrumented treadmill was used. First, the participants were asked to start running on the treadmill at 1.5 m/s and then gradually increase their speed. This helped the runners to warm up and also was used to determine the comfortable speed for each runner to perform the experiments. Eight different running styles were considered for this experiment: 
(1) Egg-Beater Gait (where instead of only moving forward, the leg also travels laterally to the side);
(2) Bouncing Gait (where the vertical bounce after each step is higher than normal); 
(3) Heel Strike (where landing occurs on the heels as opposed to mid-foot landing);
(4) Toe Strike (where landing occurs on the fore-foot);
(5) Lengthened Strides (where the strides are abnormally longer than normal);
(6) Shortened Strides (where the strides are abnormally shorter than normal); 
(7) Wide Stance (where the horizontal space between the two feet is wider than normal); and (8) Narrow Stance (where the horizontal space between the two feet is shorter than normal). These particular classes were selected as they encapsulate exaggerated common variations in running styles among people. Each subject was asked to perform these running styles for approximately 5 minutes after each style was verbally described to them. This approach allowed for subjects to perform slight variations of each style given their own comfort and physiology.
Ethics approval was secured from Queen's General Research Ethics Board.

\begin{figure*}[t!]
\centerline{\includegraphics[width=0.8\linewidth]{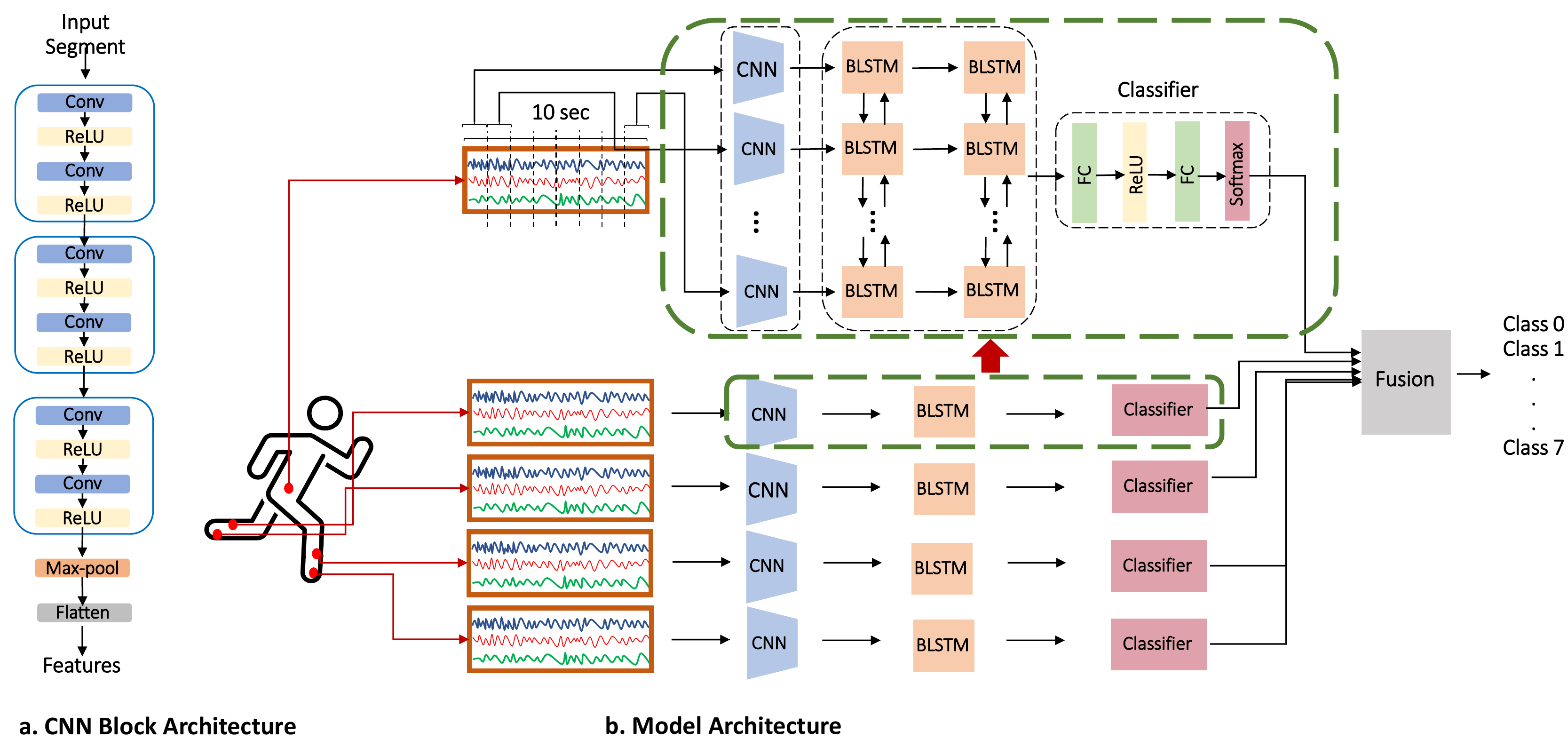}}
\caption{The architecture of the proposed CNN-LSTM model. } 
\label{fig:arch}
\end{figure*}

\subsection{Proposed Model}
In this study, we propose a CNN-LSTM architecture to classify different running styles based on input multi-sensor accelerometer data. The first part of our model is a CNN which learns to automatically extract representations from the 3-D accelerometer signals. This network consists of 3 blocks with 2 1-D convolutional layers in each block. Figure \ref{fig:arch}(a) demonstrates the CNN component of our model.
The kernel size is set to 3 and the numbers of filters for the convolutional layers in each block are 64, 128, and 256, respectively. These blocks are then followed by a 1D maxpooling layer with a pool size of 2 and a flatten layer to generate the output features from the input signals. 

Following the CNN, a Recurrent Neural Network is used in the form of a 2-layer Bidirectional LSTM (BLSTM), each consisting of 256 hidden units. The BLSTM aims to learn the temporal traits and relationships between the representations extracted by the CNN from input raw accelerometer data in both forward and backward directions. The output of the final bidirectional layer is then fed to 2 fully connected layers (where the first one is followed by ReLU activation) of sizes 200 and 8 (number of running classes) and a softmax layer. A similar CNN-LSTM architecture is used for the data from each sensor, and the final outputs of the 5 CNN-LSTM networks are fused through score-level fusion (averaging the output probabilities). Figure \ref{fig:arch}(b) presents the overall architecture of our proposed solution. The dashed green rectangle in the figure illustrates the details of the network used for each sensor.

To feed each CNN-LSTM model, we divide the running data into segments of 10 seconds in duration, with 50\% overlap. As shown in Figure \ref{fig:arch}, the input of the network is a 10-second 3-D acceleration signal collected from each of the five IMUs placed on different parts of the body. Each 10-second input signal is segmented into 8 successive sub-segments of 1.25 seconds in duration with no overlap. The 1.25 second sub-segments are fed to the CNN network to generate vector representations, to subsequently feed 8 LSTM cells organized in 2 bidirectional layers.

\begin{table*}[!t]

\caption{Results for the k-fold cross-validation scheme.}
\begin{adjustbox}{max width=1\textwidth}
\small
\begin{tabular}{l|l|l|l|l|l|l|l|l|l|l||l|l}
\hline

     \textbf{\multirow{2}{*}{Models}} & 
     \multicolumn{2}{c|}{\textbf{COM}} &
     \multicolumn{2}{c|}{\textbf{LFoot}} & \multicolumn{2}{c|}{\textbf{LShank}} & \multicolumn{2}{c|}{\textbf{RFoot}} & \multicolumn{2}{c||}{\textbf{RShank}} &
     \multicolumn{2}{c}{\textbf{Fusion}} \\ \cline{2-13}

 &  \textbf{Acc.} & \textbf{F1} &  \textbf{Acc.} & \textbf{F1} &  \textbf{Acc.} & \textbf{F1} &  \textbf{Acc.} & \textbf{F1} &  \textbf{Acc.} & \textbf{F1} &  \textbf{Acc.} & \textbf{F1} \\
\hline\hline

\textbf{Naive Bayes} &  0.309$\pm$0.012 & 0.294$\pm$0.012 &  0.306$\pm$0.016 & 0.289$\pm$0.013 &  0.323$\pm$0.015 & 0.298$\pm$0.013 &  0.316$\pm$0.019 & 0.291$\pm$0.015 &  0.332$\pm$0.014 & 0.302$\pm$0.010 &  0.317$\pm$0.018 & 0.295$\pm$0.014 \\
\textbf{Decision Tree} &  0.560$\pm$0.020 & 0.560$\pm$0.019 &  0.478$\pm$0.016 & 0.477$\pm$0.016 &  0.603$\pm$0.012 & 0.603$\pm$0.013 &  0.472$\pm$0.010 & 0.471$\pm$0.010 &  0.610$\pm$0.012 & 0.610$\pm$0.011 &  0.545$\pm$0.061 & 0.544$\pm$0.061 \\
\textbf{SVM} &  0.689$\pm$0.011 & 0.688$\pm$0.012 &  0.658$\pm$0.013 & 0.657$\pm$0.012 &  0.715$\pm$0.013 & 0.714$\pm$0.013 &  0.614$\pm$0.008 & 0.613$\pm$0.007 &  0.745$\pm$0.011 & 0.745$\pm$0.011 &  0.685$\pm$0.046 & 0.683$\pm$0.046 \\
\textbf{BTE} &  0.763$\pm$0.015 & 0.763$\pm$0.015 &  0.676$\pm$0.003 & 0.675$\pm$0.003 &  0.750$\pm$0.007 & 0.750$\pm$0.008 &  0.669$\pm$0.013 & 0.0.668$\pm$0.011 &  0.765$\pm$0.004 & 0.765$\pm$0.005 &  0.724$\pm$0.044 & 0.724$\pm$0.044 \\
\textbf{CNN} &  0.812$\pm$0.004 & 0.814$\pm$0.004 &  0.823$\pm$0.012 & 0.826$\pm$0.013 &  0.817$\pm$0.010 & 0.820$\pm$0.007 &  0.834$\pm$0.011 & 0.839$\pm$0.012 &  0.813$\pm$0.015 & 0.815$\pm$0.015 &  0.863$\pm$0.010 & 0.865$\pm$0.010 \\
\hline
\textbf{CNN-LSTM} &  0.857$\pm$0.010 & 0.859$\pm$0.009 &  0.865$\pm$0.005 & 0.866$\pm$0.004 &  0.860$\pm$0.007 & 0.861$\pm$0.007 &  0.870$\pm$0.014 & 0.871$\pm$0.014 &  0.870$\pm$0.010 & 0.873$\pm$0.010 &  0.931$\pm$0.008 & 0.932$\pm$0.007 \\ \hline
    
\end{tabular}
\end{adjustbox}

\centering
\label{tab:kfold}
\end{table*}

\begin{table*}[!t]
\caption{Results for the LOSO scheme. FT indicated fine-tuning with 5\% of unseen subjects' data.}
\begin{adjustbox}{max width=1\textwidth}
\small
\begin{tabular}{l|l|l|l|l|l|l|l|l|l|l||l|l}
\hline
 
     \textbf{\multirow{2}{*}{Models }} & 
     \multicolumn{2}{c|}{\textbf{COM}} &
     \multicolumn{2}{c|}{\textbf{LFoot}} & \multicolumn{2}{c|}{\textbf{LShank}} & \multicolumn{2}{c|}{\textbf{RFoot}} & \multicolumn{2}{c||}{\textbf{RShank}} &
     \multicolumn{2}{c}{\textbf{Fusion}} \\ \cline{2-13}

 &  \textbf{Acc.} & \textbf{F1} &  \textbf{Acc.} & \textbf{F1} &  \textbf{Acc.} & \textbf{F1} &  \textbf{Acc.} & \textbf{F1} &  \textbf{Acc.} & \textbf{F1} &  \textbf{Acc.} & \textbf{F1} \\
\hline\hline

\textbf{Naive Bayes} &  0.144$\pm$0.035 & 0.103$\pm$0.032 &  0.172$\pm$0.029 & 0.118$\pm$0.038 &  0.183$\pm$0.043 & 0.129$\pm$0.032 &  0.259$\pm$0.0.118 &0.214$\pm$0.134 &  0.287$\pm$0.169 & 0.262$\pm$0.191 &  0.209$\pm$0.027 & 0.165$\pm$0.043 \\

\textbf{Decision Tree} &  0.140$\pm$0.033 & 0.126$\pm$0.031 & 0.150$\pm$0.031 &  0.139$\pm$0.030 & 0.168$\pm$0.047 &  0.148$\pm$0.042 & 0.423$\pm$0.324 &  0.414$\pm$0.331 & 0.452$\pm$0.352 &  0.440$\pm$0.362 &  0.267$\pm$0.093 & 0.253$\pm$0.096 \\

\textbf{SVM} &  0.187$\pm$0.028 & 0.168$\pm$0.031 &  0.196$\pm$0.029 & 0.174$\pm$0.027 &  0.207$\pm$0.084 & 0.193$\pm$0.077 &  0.491$\pm$0.358 & 0.475$\pm$0.371 &  0.440$\pm$0.283 & 0.421$\pm$0.294 &  0.304$\pm$0.086 & 0.286$\pm$0.093 \\

\textbf{BTE} &  0.193$\pm$0.029 & 0.167$\pm$0.028 &  0.184$\pm$0.052 & 0.167$\pm$0.050 &  0.192$\pm$0.035 & 0.169$\pm$0.039 &  0.520$\pm$0.393 & 0.502$\pm$0.407 &  0.525$\pm$0.389 & 0.507$\pm$0.402 &  0.323$\pm$0.100 & 0.303$\pm$0.106 \\

\textbf{CNN} &  0.259$\pm$0.062 & 0.264$\pm$0.055 &  0.347$\pm$0.085 & 0.351$\pm$0.082 &  0.408$\pm$0.111 & 0.443$\pm$0.102 &  0.298$\pm$0.114 & 0.290$\pm$0.105 &  0.339$\pm$0.111 & 0.356$\pm$0.105 &  0.376$\pm$0.083 & 0.401$\pm$0.063 \\
\textbf{CNN (FT)} & 0.475$\pm$0.059 & 0.475$\pm$0.057 &  0.484$\pm$0.096 & 0.491$\pm$0.091 &  0.489$\pm$0.061 & 0.493$\pm$0.057 &  0.479$\pm$0.040 & 0.519$\pm$0.047 &  0.530$\pm$0.091 & 0.578$\pm$0.084 &  0.618$\pm$0.067 & 0.654$\pm$0.067 \\
\hline
\textbf{CNN-LSTM} &  0.265$\pm$0.044 & 0.276$\pm$0.054 &  0.365$\pm$0.084 & 0.372$\pm$0.091 &  0.374$\pm$0.095 & 0.378$\pm$0.105 &  0.293$\pm$0.029 &  0.301$\pm$0.031 &  0.337$\pm$0.094 & 0.360$\pm$0.091 &  0.385$\pm$0.045 & 0.404$\pm$0.038 \\ 
\textbf{CNN-LSTM (FT)} &  0.608$\pm$0.086 & 0.638$\pm$0.084 &  0.609$\pm$0.080 & 0.637$\pm$0.081 &  0.636$\pm$0.072 & 0.663$\pm$0.063 &  0.585$\pm$0.058 &  0.615$\pm$0.062 &  0.578$\pm$0.084 & 0.607$\pm$0.080 &  0.714$\pm$0.053 & 0.746$\pm$0.052 \\
\hline
    
\end{tabular}
\end{adjustbox}
\centering
\label{tab:loso}
\end{table*}

\begin{table}[!t]
\centerline{}
\begin{center}
\caption{Results for fine-tuning the proposed CNN-LSTM model with LOSO.}
\begin{adjustbox}{max width=0.7\textwidth}
\scriptsize
\begin{tabular}{lll}
\hline
\textbf{Tuning} & \textbf{Acc.} & \textbf{F1} \\ \hline\hline
\textbf{No Tuning}        &0.385$\pm$0.045          &0.404$\pm$0.038      \\
\textbf{Tuning 2\%}        &0.596$\pm$0.084          &0.637$\pm$0.076      \\
\textbf{Tuning 5\%}        & 0.714$\pm$0.053           &0.746$\pm$0.052    \\
\textbf{Tuning 10\%}       & 0.766$\pm$0.050           &0.789$\pm$0.047    \\
\textbf{Tuning 20\%}        &0.8067$\pm$0.041          &0.825$\pm$0.040    \\ 
\hline     
\end{tabular}
\end{adjustbox}
\label{tab:Tunning}
\end{center}
\end{table}

\subsection{Implementation and Evaluation}
\label{subsec:Implementation Details and Evaluation}

We implement our proposed model using Keras with TensorFlow backend on an NVIDIA GeForce RTX 2080 Ti GPU. The models are trained over 300 epochs with a batch size 64 and Adam optimizer with a learning rate of $0.0002$. 

In order to evaluate our method, we use 2 schemes: k-folds cross-validation and Leave-One-Subject-Out (LOSO). For k-folds, we randomly picked 20\% of the segments for testing, while using the rest for training. We carried out this approach for 5 trials and presented the means and standard deviations of accuracy and F1 scores among all 5 folds. In each trial, 10\% of the training set was allocated for validation. In LOSO, for each trial, 20\% of the subjects were selected for testing and the remaining subjects were set to train the model. Similar to the previous approach, 10\% of the training data was set aside for validation.

\subsection{Baselines}

In order to better evaluate the robustness of our proposed method, several common classification methods are applied for comparison. Specifically, 4 classical machine learning classifiers are used. These classifiers are: Naive Bayes, Decision Tree, SVM, and Bagged Tree Ensemble (BTE). We extract 8 common manual features (mean, standard deviation, minimum, maximum, root mean square, skewness, kurtosis, and peak-to-peak time) from each segment to train these classical models.

The hyperparameters for each model are tuned with the help of Bayesian optimization. For SVM, cubic kernel and one-vs-one scheme were the best fits for all the sensors, with the exception of the LFoot sensor where a Gaussian kernel function and one-vs-all scheme showed the best results. For the Decision Tree, maximum deviance reduction was set as the Split criterion for all sensors except for the COM sensor which used the Towing rule. 


In addition to the 4 classical machine learning methods, a CNN was designed to automatically extract the features from each input signals and classify the running styles based on the 5 wearable accelerometers. This also forms an ablated version of our model, where the BLSTM is removed. Successive to removal of the BLSTM network, the hyperparameters of the CNN were further optimized to yeild the best results. The CNN baseline consists of 4 convolution blocks with 128, 256, 384, and 512 filters, followed by 4 FC layers with 384, 200, 120, and 8 hidden neurons. For all the convolution layers, the kernel size was set to 3. The models were trained over 300 epochs with a batch size 64 and Adam optimizer with a learning rate of $0.0002$.

\section{Results}
\begin{figure}[t!]
    \centering
    \includegraphics[width=0.7\columnwidth]{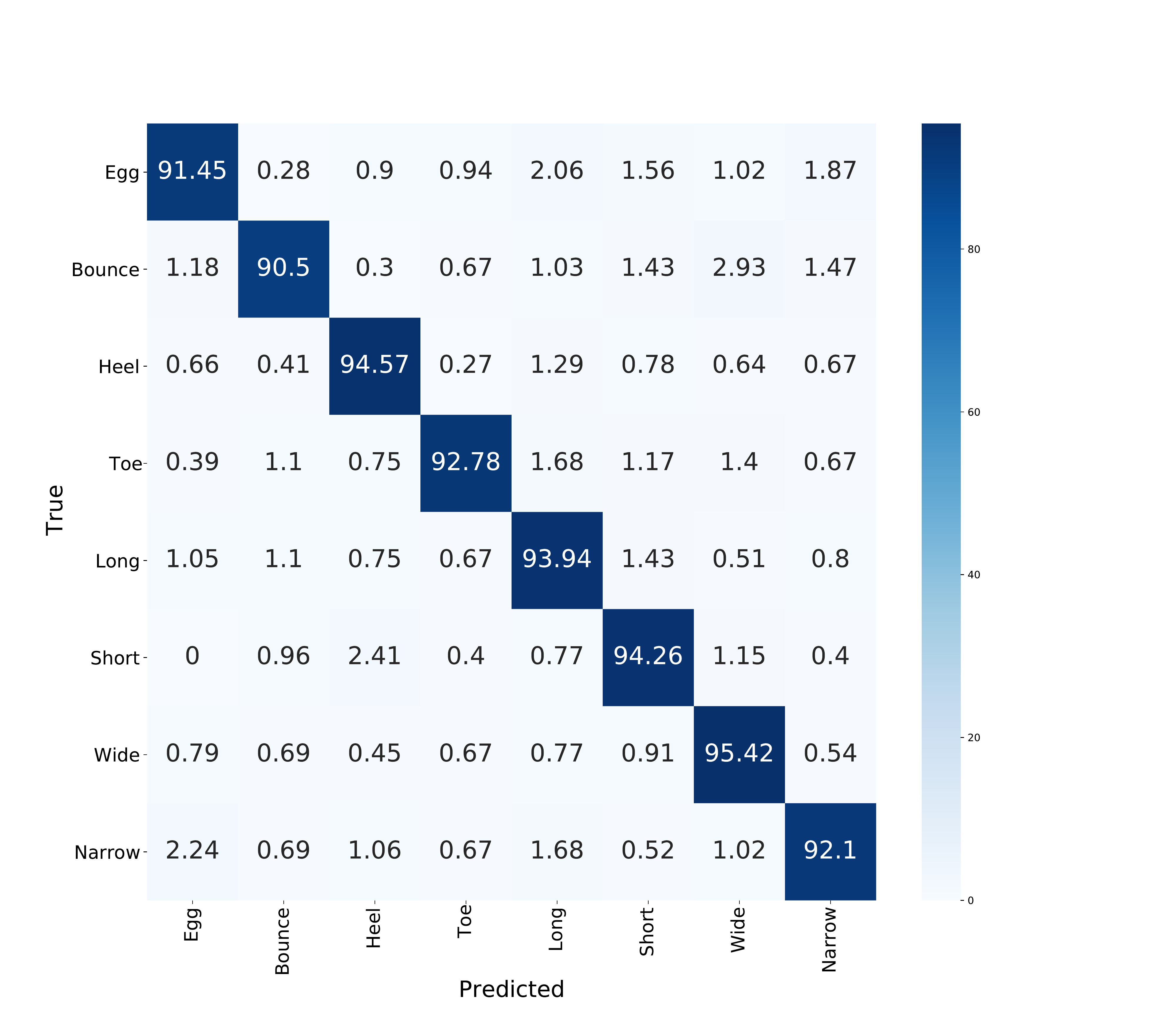}
    \caption{Confusion matrix for the proposed method.}
    \label{fig:Confusion Matrix for CNN-LSTM in the kfold split scheme}
\end{figure}

In this section, we present the results of our experiments described in the previous section. Our proposed CNN-LSTM method outperforms all the baselines in both k-fold and LOSO schemes. Table \ref{tab:kfold} presents the accuracy and F1 scores for all the models with k-fold cross-validation. It can be observed that the proposed CNN-LSTM model obtains the best results achieving an accuracy and F1-score of 0.931$\pm$0.008 and 0.932$\pm$0.007 respectively. Among the baseline models, the CNN shows the best performance, outperforming the classical machine learning solutions. Exploring the performance of the model on different parts of the body (sensor locations) indicates that all the sensors provide effective information for classification of running styles. Moreover, we observe that by fusing the 5 sensors pipelines, a boost in performance is achieved, demonstrating that different parts of the body contain complimentary information that can be exploited for automatic classification of running styles. We also present the confusion matrix for this model in Figure \ref{fig:Confusion Matrix for CNN-LSTM in the kfold split scheme}, where we observe that different running styles have been classified consistently across the board. Meanwhile, the `wide-stance' style proves the easiest to classify whereas the `bouncing' style proves the most difficult.

The results of the models in the LOSO scheme are presented in Table \ref{tab:loso}. It can be observed that while our proposed CNN-LSTM model outperforms the baselines, the overall performance of all the models, both baselines and the proposed, significantly drop. It can therefore be inferred that variations in running styles are highly subject-specific, making it difficult for automatic subject-independent classification, highlighting the need for solutions capable of learning generalized representations via transfer learning or self-supervised learning. Moreover, the fact that subjects performed the running styles based on their comfort could be another contributing factor to this matter. To further evaluate this matter, we fine-tune both CNN and CNN-LSTM models with a small fraction of test data (5\%). As observed in Table \ref{tab:loso}, for both models, the performance is significantly improved, reaching an accuracy of 0.618 for CNN model and 0.714 for CNN-LSTM model. The results of fine-tuning the proposed CNN-LSTM model with various portions of test data are presented in Table \ref{tab:Tunning}. As expected, increasing the amount of data used for fine-tuning improves the performance of the model, while we observe that utilizing even a very small amount of user-specific data (2\%), results in a significant boost in performance.

\section{Conclusions and Future Work}
In this study, we implemented a system for classifying running style variations. For this purpose, we collected data using 5 wearable accelerometers from 10 healthy runners while performing 8 different running styles on a treadmill. We then proposed a deep CNN-LSTM architecture to detect these target classes by learning to automatically extract effective representations followed by learning temporal dependencies in the time-series. We compared the performance of our solution to a number of baseline techniques, where CNN-LSTM model showed the best results. 
Our findings suggest that even pre-defined running styles are personalized for different people and based on comfort and physiology, making it difficult for machine learning and deep learning methods to automatically classify them, however fine-tuning the model with a small portion of each subject's samples, significantly improved the LOSO performance.

For future work, we will explore approaches such as transfer learning, self-supervised learning, and contrastive learning, that have proven effective in enabling deep learning solutions to extract highly generalized features from input data. Moreover, domain transfer techniques may be explored to perform cross-subject domain adaptation.



\small
\bibliographystyle{unsrt}
\bibliography{b.bib}

\end{document}